# An evaluation of local shape descriptors for 3D shape retrieval

Sarah Tang[a], Afzal Godil[b]
[a]Princeton University, Princeton, NJ, USA;
[b]National Institute of Standards and Technology, Gaithersburg, MD , USA

## ABSTRACT

As the usage of 3D models increases, so does the importance of developing accurate 3D shape retrieval algorithms. A common approach is to calculate a shape descriptor for each object, which can then be compared to determine two objects' similarity. However, these descriptors are often evaluated independently and on different datasets, making them difficult to compare. Using the SHREC 2011 Shape Retrieval Contest of Non-rigid 3D Watertight Meshes dataset, we systematically evaluate a collection of local shape descriptors. We apply each descriptor to the bag-of-words paradigm and assess the effects of varying the dictionary's size and the number of sample points. In addition, several salient point detection methods are used to choose sample points; these methods are compared to each other and to random selection. Finally, information from two local descriptors is combined in two ways and changes in performance are investigated. This paper presents results of these experiments.

## 1. INTRODUCTION

Collections of 3D models are becoming prominent in many fields. As the number and sizes of such databases grow, so does the need to accurately search for and retrieve 3D models. Thus, a standard 3D shape retrieval problem becomes returning all other objects in the database ordered in decreasing similarity to a given query.

To approach this problem, a number of shape descriptors have been proposed. Global descriptors, such as area and volume [1], "shape distributions" [2], ratios derived from the object's convex-hull [3], or the "LightField Descriptor" [4], return a feature vector, a single vector of values to represent an object. The distance between two feature vectors then quantitatively represents the dissimilarity between their corresponding objects; the more similar the objects are, the lower their dissimilarity value, and a value of 0 indicates the objects are identical. However, global descriptors are often not invariant to scaling, rotation, or translation, and can only discriminate between broad categories. Thus, local descriptors, which use a single vector to describe the local surface region around a number of sample points on an object, are sometimes used instead.

The bag-of-words paradigm offers a framework with which to compare two objects using local descriptors. Briefly, for each object, a selected set of sample points is each associated with a local descriptor value from a pre-constructed dictionary, or a visual word. A feature vector for that object then has dimensionality equal to the size of the dictionary; its values are the histogram counts for the number of occurrences of each visual word. Objects can be compared with a dissimilarity value, just as in the case of global descriptors.

This method has been successful in both text and image retrieval, and has shown promising results in 3D. However, while many local descriptors have been proposed, only a few have been incorporated into the bag-of-words framework. Many times, descriptors will be incorporated into different algorithms and tested on different datasets. This paper systematically surveys a set of local descriptors and compares their performances. It first analyzes their individual performances. In addition, parameters for the bag-of-words algorithm, four different salient point detection methods, and ways to combine information from two local descriptors are investigated.

This paper will proceed as follows: Section 2 briefly discusses related work. Section 3 gives an overview of the bag-of-words approach, Section 4 describes our experiments, and Section 5 outlines the evaluation methods. Section 6 presents our results, while Section 7 concludes the paper.

## 2. RELATED WORK

Some works have already described methods using the bag-of-words approach. Ohbuchi et al. [5] devise a bag-of-features based on Scale-Invariant Feature Transform (SIFT) algorithm, where SIFT is used to extract salient local features from uniformly distributed depth-buffer views of a normalized object. The SIFT features from all views are used to construct a dictionary and create a histogram; dissimilarity is the Kullback-Leibler divergence between two histograms. Lian et al. [6] use a similar method, but instead of constructing one global histogram for each object, a separate histogram is built for each view of the object. They then compare objects with a method called "Clock Matching". Shan et al. [7] and Liu et al. [8] both use spin images, 2D histograms proposed by Johnson and Herbert [9] that count the number of surface points at various locations, as local descriptors. Toldo et al. [10] segment a model into regions, calculate different sets of "region descriptors", and apply a "multi-clustering approach". Lavoué [11] propose choosing a random set of vertices as seeds and applying Lloyd relaxation iterations to create a uniform sampling. They then define a local descriptor to describe each sample point's local surface patch.

Additionally, a number of local descriptors have been proposed independent of the bag-of-words framework. Curvature based local descriptors include mean and Gaussian curvature, shape index [12], and curvedness [12]. Sun et al. [13] introduce the popular Heat Kernel Signature (HKS). Gal et al. [14] create a global 2D histogram that combines two local descriptors – the local-diameter function, which describes local shape well, and the centricity function, which gives spatial information.

With these local descriptors comes the challenge of selecting appropriate and repeatable sample points. Some successful 2D salient point detection methods have been extended to 3D, such as with Sipiran and Bustos's 3D-Harris [15]. Multi-scale approaches are also common; for example, [16] and [17] calculate saliency measures for each vertex using Gaussian filters with different standard deviations and choose certain maximum values. [18] define an "integral volume descriptor", which calculates the volume of the object's interior contained by the intersection between its surface and a ball centered on one of its vertices. The Heat Kernel Signature proposed by Sun et al. [13] can also be used to find salient points.

A number of surveys have been conducted to compare these methods under one framework. Tangelder and Veltkamp [19] review content-based 3D shape retrieval methods. Heider et al. [20] assess local descriptors for stability and discrimination ability, while Dutagaci et al. [21] examine salient point detection methods based on human-established ground truth. Li and Godil [22] evaluate the bag-of-words method in context of different shape retrieval tasks, choosing spin images as the local descriptor. This evaluation contributes by exploring the performance of different local shape descriptors, specially applied to the bag-of-words method.

## 3. BAG-OF-WORDS

The bag-of-words approach involves two stages: building a visual dictionary and computing a corresponding histogram for each object. Figure 1 illustrates the algorithm.

To construct a visual dictionary, we choose $\{n_1, n_2, n_3...n_M\}$ sample points from a collection of *M* models. A local descriptor is evaluated at each point, resulting in a total of $\sum_{i-1}^{M} n_i$ vectors. We then cluster these vectors into *D* clusters, where *D* is the desired size of the dictionary, and the cluster centers are designated as the visual words.

For an object *i*, we choose $n_i$ sample points and evaluate the local descriptor at these points. Each resulting vector is associated with the nearest word in the visual dictionary, and a histogram counting the number of occurrences of each visual word is constructed.

In our implementation, we choose to base the visual dictionary and the histograms off the same set of sample points. We use code by Vedaldi and Fulkerson [23] to implement *k*-means clustering. Since we do not always choose the same number of sample points for each object, we divide counts by the total number of sample points to normalize histograms to [0, 1]. We choose Euclidian distance as a distance measure. For a dataset containing *M* meshes, we create an *M*x*M* distance matrix, where the value at entry (*i, j*) is the distance between the histograms of objects *i* and *j* and consequently, their dissimilarity value.

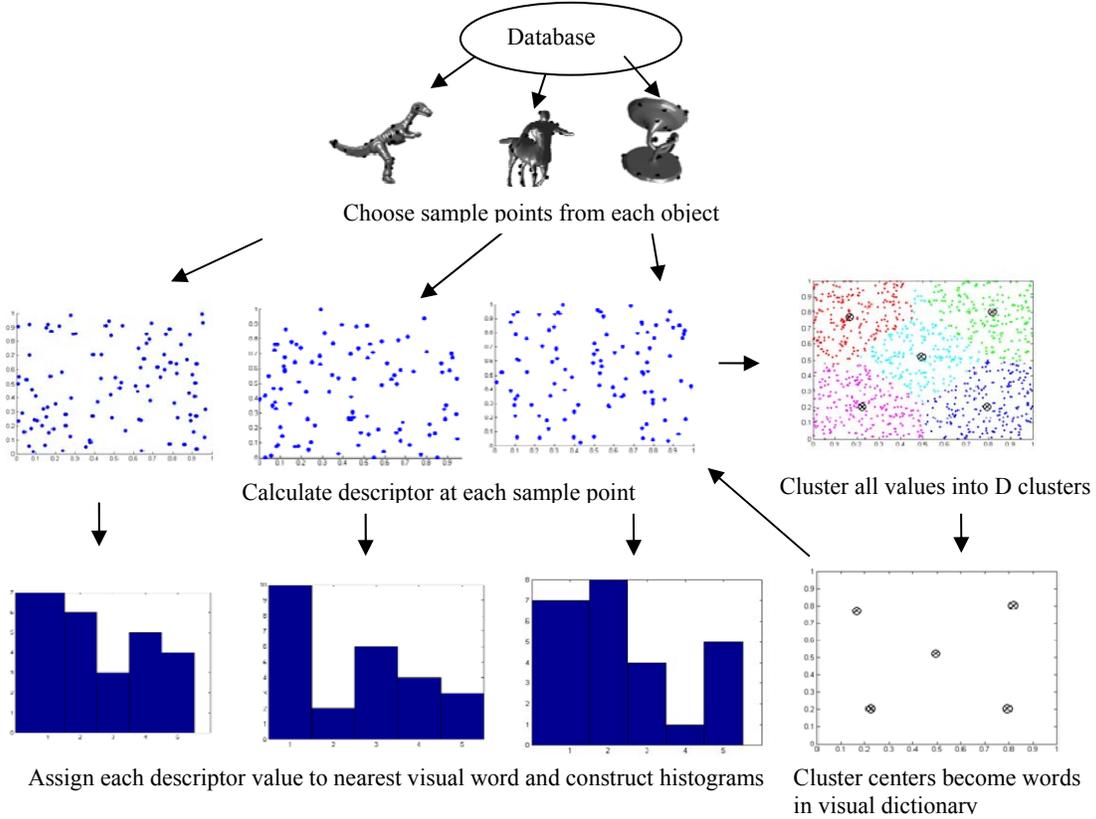

Figure 1. The bag-of-words algorithm.

## 4. EXPERIMENTS

For our experiments, we assume that all objects are modeled as triangular meshes, where the surface of an object is approximated as a set of edge-connected triangles defined by a set of vertices and edges.

### 4.1 Local descriptors

We adopt the method of Heider et al. [20] to evaluate local descriptors for our dataset. The procedure is outlined as follows:

**Calculation method**

Descriptors are sampled across a small area surrounding each vertex $P$. Sample points $Q$ are taken in intervals of $2\pi r/20$ along each of $R$ rings formed by the intersections of $R$ spheres, with center $P$ and varying radius $r$, with the object's surface. Sampling is done with what [20] denote as histogram sampling – within each ring, the sampled values are sorted and the ones at 0%, 10%, 30%, 50%, 70%, 90%, and 100% are taken as sample points. Thus, each point $P$ is associated with a vector of size $RS$, where $R$ is the number of spheres and consequently, the number of rings and $S$ is the number of samples kept per ring (in this case, $S = 7$).

For all descriptors, $R = 5$ and $r$ is determined with the equation
$$r = B0.0375/R, \qquad (1)$$
where $B$ is the diagonal of the surface's bounding box. Descriptors are sampled at all vertices, associating an object containing $V$ vertices with $V$ vectors, each with dimensionality $RS$.

PCA is applied across vectors from all vertices on all objects in the dataset to keep data comparable after reduction. After the eigenvalues drop below 10 % of the largest value, remaining coefficients are dropped. All vectors are re-projected onto this new coordinate system and values in each dimension are normalized to [0, 1].

**Descriptors**

We test the six local descriptors that yielded the best performance in the evaluation by Heider et al. [20]. They are as follows:

- Distance to plane [DTP]: the descriptor is the signed distance from each sample point on the ring, $Q$, to the best-fit plane of that ring
- Normal distribution [ND]: the descriptor contains two values – one is the angle between the vertex $P$'s normal and the projection of the sample point $Q$'s normal onto the best-fit plane to $P$, $P$'s normal, and $Q$, the other is the projection of $Q$'s normal onto the best-fit plane of $Q$, $Q$'s normal, and an adjacent sample point
- Mean curvature [Mean]: $\frac{\kappa_1+\kappa_2}{2}$, where $\kappa_1$ and $\kappa_2$ are the principal curvatures at point $P$
- Gaussian curvature [Gauss]: $\kappa_1\kappa_2$, where $\kappa_1$ and $\kappa_2$ are the principal curvatures at point $P$
- Shape index [SI]: calculated with

$$\frac{2}{\pi}\tan^{-1}\left(\frac{\kappa_2+\kappa_1}{\kappa_2-\kappa_1}\right), \qquad (2)$$

  where $\kappa_1$ and $\kappa_2$ are the principal curvatures at point $P$ and $\kappa_1 \geq \kappa_2$
- Curvature index [CI]: calculated with

$$\sqrt{\frac{\kappa_1^2+\kappa_2^2}{2}}, \qquad (3)$$

  where $\kappa_1$ and $\kappa_2$ are the principal curvatures at point $P$

Local descriptors are calculated with code provided by the authors [24].

### 4.2 Choice of sample points

We compare different sampling methods – choosing a uniform random sampling of vertices and four different salient point detection methods. We wonder if by sampling local descriptors only at important points, we can either improve or maintain performance while choosing fewer sample points. The point detection methods tested are:

**Mesh saliency**

Proposed by Lee et al. [16], points are chosen according to their "total saliency" values. "Saliency maps", denoted $\gamma_i$, are created by calculating a saliency value for each vertex. For a chosen scale $i$, the Gaussian-weighted averages of the mean curvature using two Gaussian filters, with standard deviations $\sigma_i$ and $2\sigma_i$, are calculated. The absolute difference between the two averages is the saliency. "Saliency maps" are calculated at five scales, with $\sigma_i = \{2\varepsilon, 3\varepsilon, 4\varepsilon, 5\varepsilon, 6\varepsilon\}$, where $\varepsilon$ is 0.3 % of the diagonal of the model's bounding box. Each "saliency map" is normalized, and the five are combined with a non-linear suppression operator $S(x)$ to find the "total saliency" $\gamma$ using the equation

$$\gamma = \sum_i S(\gamma_i). \qquad (4)$$

Vertices whose "total saliency" values are local maxima, higher than all its neighboring vertices' values, are identified as candidate points. All candidate points whose "total saliency" is higher than the average "total saliency" of all local maxima values become salient points. Code implemented by Dutagaci et al. [21] is used for our experiment.

**Salient points**

Proposed by Castellani et al. [17], salient points are determined using a "joint multi-scale" paradigm. The algorithm is split into an intra-octave and an inter-octave phase. To find intra-octave salient points for the object re-meshed at decimation $d$, denoted $M^d$ and chosen scale $i$, the Difference-of-Gaussians operator, $F_i^d(P)$ calculated for a vertex $P$ by finding the difference between Gaussian operators with standard deviations $\sigma_i$ and $2\sigma_i$. $F_i^d(P)$ is projected onto the normal of $P$ to obtain a "scale map", $M_i^d(P)$. "Scale maps" are normalized, and an "inhibited scale map", $\widehat{M}_i^d(P)$ – where for each vertex, $\widehat{M}_i^d(P) = M_i^d(P)$ only if its value is higher than 85 % of values in its neighborhood, otherwise, $\widehat{M}_i^d(P) = 0$ – is formed and added to $M_i^d(V)$ to get an "inhibited saliency map." This is done at six scales, with $\sigma_i = \{\varepsilon, 2\varepsilon, 3\varepsilon, 4\varepsilon, 5\varepsilon, 6\varepsilon\}$, where $\varepsilon$ of 0.1 % the main diagonal of the model's bounding box. A "non-maximum suppression" scheme detects salient points for the mesh at decimation $M^d$.

This intra-octave phase is carried out on meshes $M^d$ with $d = \{0, h, 2h, 3h, 4h\}$, where $h = 0.20$. Points common to three of these meshes become final salient points. The "Mesh Tool" program is used for point detection [25].

**3D-Harris adaptive**

Sipiran and Bustos detect salient points by extending 2D corner detection to 3D models [15]. The neighborhood $N_k(P)$ of vertex $P$ is defined as the $k$ rings around $P$. The centroid of $N_r(P)$ becomes the origin of the 3D coordinate system. PCA is applied to $P$ and its neighborhood to compute a best-fit plane, and points are re-oriented such that $P$ becomes the origin and the best-fit plane becomes the $xy$-plane of a transformed coordinate system. A quadratic surface is fit to the transformed points and its derivatives, $f_x$ and $f_y$, are calculated. A matrix $E$ is computer with:

$$E = \begin{bmatrix} \frac{1}{\sqrt{2\pi\sigma}} \int_{R^2} e^{\frac{-(x^2+y^2)}{2\sigma^2}} \cdot f_x(x,y)^2 dxdy & \frac{1}{\sqrt{2\pi\sigma}} \int_{R^2} e^{\frac{-(x^2+y^2)}{2\sigma^2}} \cdot f_x(x,y) \cdot f_y(x,y) dxdy \\ \frac{1}{\sqrt{2\pi\sigma}} \int_{R^2} e^{\frac{-(x^2+y^2)}{2\sigma^2}} \cdot f_x(x,y) \cdot f_y(x,y) dxdy & \frac{1}{\sqrt{2\pi\sigma}} \int_{R^2} e^{\frac{-(x^2+y^2)}{2\sigma^2}} \cdot f_y(x,y)^2 dxdy \end{bmatrix}. \quad (5)$$

The Harris operator $H(V)$ is calculated with
$$H(V) = \det(E) - k.(tr(E))^2, \quad (6)$$
where $E$ is the matrix defined in (5), and $k$ is a chosen constant referred to as the Harris parameter. Points are considered salient if they are local maxima for its neighborhood of $r$ rings. In addition, a specified fraction of points with the overall highest Harris response values are also chosen.

C++ code provided by the authors is used for evaluation [26]. In the default implementation, an adaptive sampling scheme finds an appropriate neighborhood $n$ for each vertex $P$ depending on $P$'s surrounding tessellation, selecting a different neighborhood for different points. 1 % of the diagonal of the object's bounding box is used as a parameter to find this neighborhood. The Harris parameter $k$ is 0.04, the neighborhood size for determining saliency $r$ is 1, and 1 % of vertices are selected as salient points. We refer to this method as "3D-Harris adaptive".

**3D-Harris rings**

In addition to the default implementation of 3D-Harris, sample points chosen with different parameter values are tested with our bag-of-words approach. The parameters altered are, as outlined at the author's website [26]:

- Neighborhood type: the method used – adaptive or rings – to find each vertex's neighborhood. 3D-Harris adaptive uses the adaptive scheme. With the rings option, the neighborhood is simply a constant number of rings around a vertex.
- Parameter-neighborhood: the fraction of the diagonal of the object's bounding box if neighborhood type is adaptive; the number of rings used if neighborhood type is rings
- $k$: constant in (6)
- Ring-maxima-detection: number of rings considered as the neighborhood when finding local maxima
- Parameter-selection: the fraction of the total number of vertices selected as salient points

The salient point selection method can alternatively be a clustering technique, where points are sorted in order of descending Harris operator values and clustered before salient points are selected. Because this technique intends to get an even distribution of interest points, it yields sample points visually similar to a random distribution. Thus, we choose to always select a fraction of points. When varying a parameter does not affect performance, the lowest value is chosen. The parameters returning points giving the best performance is used for the "3D-Harris rings" method.

### 4.3 Multiple local descriptors

Finally, we determine whether using two local descriptors would cause improved performance. Two methods are tested:

**Concatenating vectors**

Given two local descriptors that return vectors of sizes $RS_1$ and $RS_2$, at each point $P$, the second descriptor's vector is concatenated to the end of the first to create a vector of size $R(S_1+S_2)$. With this new set of vectors, clustering and histogram construction is performed as before.

**Concatenating histograms**

Two local descriptors are calculated at each point $P$. Clustering is performed independently on both sets of vectors, creating two dictionaries, each of size $D$, and corresponding histograms are independently constructed and normalized. Thus, each point $P$ is associated with two histograms; these histograms are then concatenated to yield a histogram of $2D$ elements, and dissimilarity is the distance between the concatenated histograms.

## 5. EVALUATION

### 5.1 Database

We test our algorithms on the SHREC 2011 Shape Retrieval Contest of Non-rigid 3D Watertight Meshes dataset [27]. The collection contains 600 non-rigid 3D triangle meshes, classified into 30 different categories each containing 20 objects. Given these models, our algorithm computes a 600x600 distance matrix, where the value at entry $(i, j)$ is the computed dissimilarity between objects $i$ and $j$. From this, the ranked list of similar objects can be derived for any query.

### 5.2 Performance measures

For each distance matrix output, six statistics, as described in [28], are calculated to evaluate the method. For any query object $M$, let $C$ be the number of objects in $M$'s class (including itself) and $K$ the number of closest matches examined. The statistics calculated are:

- Precision-recall plot: Precision represents the portion of the $K$ closest matches returned that are in the correct class. Recall represents the portion of objects in $M$'s class that are in the top $K$ matches. Recall is plotted on the horizontal axis, while precision is plotted on the vertical axis; results closer to the horizontal line $y = 1$ are desirable.
- Nearest neighbor (NN): the mean percentage of objects in the query object's class that are also in the top $K = 1$ results; results closer to 1.0 are desirable.
- First tier (1-tier): the mean percentage of objects in the query object's class that are also in the top $K = |C| - 1$ results; results closer to 1.0 are desirable.
- Second tier (2-tier): the mean percentage of objects in the query object's class that are also in the top $K = 2 * (|C| - 1)$ results; results closer to 1.0 are desirable.
- $E$-measure: $E$-measure is defined as
$$E = 1 - \frac{2}{\frac{1}{P}+\frac{1}{R}}, \tag{7}$$
where $P$ is the precision and $R$ is the recall over the top $K = 32$ objects. Results closer to 1.0 are desirable.
- Discounted Cumulative Gain (DCG): DCG reflects the performance of the algorithm when correct results that are retrieved earlier are weighted higher than those retrieved later, and is calculated as:
$$DCG = \frac{DCG_k}{1+\sum_{j=2}^{|C|}\frac{1}{\lg_2(j)}}, \tag{8}$$
where $k$ is the total number of models in the database and $DCG_i$ is calculated as:

$$DCG_i = \begin{cases} G_i, & i = 1 \\ DCG_{i-1} + \frac{G_i}{lg_2(i)}, & otherwise \end{cases}$$ (9)

$G_i$ has a value 1 if the object that is the *i*th closest match to the query object is in the query object's class, and is 0 otherwise. Results closer to 1.0 are desirable.

Code provided by Shilane et. al [29] is used for evaluation.

## 6. RESULTS AND DISCUSSION

### 6.1 Shape descriptors

To compare individual shape descriptors, we chose 500 uniform random points from each of the 600 meshes in the SHREC 2011 dataset. We clustered the 300,000 resulting vectors into a dictionary of 500 words. Each descriptor was evaluated at the same points on each mesh.

Table 1 shows statistics for all shape descriptors, and Fig. 2 presents their precision-recall graphs.

Table 1. Retrieval statistics for local shape descriptors. 500 random points were sampled on each object and the visual dictionary had 500 words.

|  | NN | 1-tier | 2-tier | E-measure | DCG |
|---|---|---|---|---|---|
| Mean curvature | 0.9833 | 0.7448 | 0.8450 | 0.6178 | 0.9288 |
| Curvature index | 0.9733 | 0.7405 | 0.8475 | 0.6173 | 0.9231 |
| Shape index | 0.9700 | 0.7441 | 0.8733 | 0.6335 | 0.9336 |
| Normal distribution | 0.9650 | 0.7360 | 0.8461 | 0.6155 | 0.9176 |
| Gaussian curvature | 0.9367 | 0.6474 | 0.7732 | 0.5580 | 0.8840 |
| Distance to plane | 0.9000 | 0.6073 | 0.7616 | 0.5422 | 0.8629 |

It is clear that distance to plane performed the worst, with Gaussian curvature as the second worst. However, no descriptor consistently performed better than the others. Mean curvature had the highest statistics, however, never had the highest precision. On the other hand, while shape index had the highest precision for most recall values, it had the lowest precision at high recall and only the third highest statistics. Overall, the best descriptors were mean curvature, shape index, and curvature index.

Heider et al. [20] tested sampling methods in addition to local descriptors. Looking only at their overall sensitivity results for histogram sampling, their local descriptor ranking is similar to Table 1 – mean curvature, curvature index, shape index, Gaussian curvature, normal distribution, and distance to plane. However, the overall conclusions of [20] are slightly different. For example, [20] reported mean curvature and normal distribution, followed by distance to plane and Gaussian curvature, as the most discriminative descriptors. While mean curvature performed well within the bag-of-words method as well, the other three actually performed the worst. Overall, [20] found normal distribution, being one of the most discriminative and stable, to be the best overall descriptor, yet in the bag-of-words paradigm, it did not perform as well as mean curvature, shape index, or curvature index. This suggests that the performance of a descriptor under other evaluations is not necessarily indicative of its performance within the bag-of-words framework.

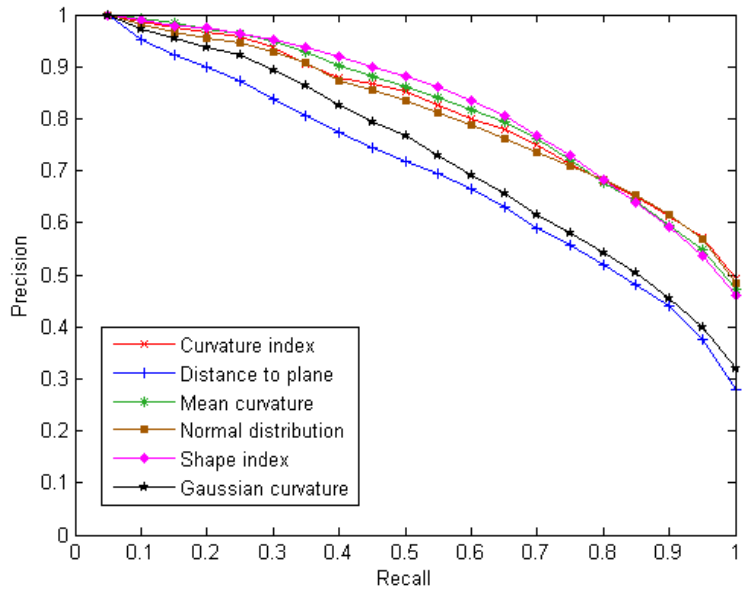

Figure 2. Precision vs. recall curves for local shape descriptors. 500 random points were sampled on each object and the visual dictionary had 500 words.

### 6.2 Algorithm parameters

We tested two important parameters of the bag-of-words method. First, we evaluated the effect of dictionary size. Choosing mean curvature as the local descriptor, we sampled 500 random points from each object. Using this same set of sample points for all tests, we created dictionaries of various sizes.

Clustering was the most time consuming step in our algorithm. It took longer for larger dictionaries, taking about 30 minutes for 1000 words.

Table 2. Time consumed for *k*-means clustering for various sized dictionaries. Timing studies done on an 2.66GHz desktop computer with 4GB of RAM running Windows XP.

| Vocabulary | Time for *k*-means clustering (seconds) |
|---|---|
| 10 | 8.52 |
| 50 | 42.09 |
| 100 | 68.85 |
| 200 | 172.24 |
| 300 | 317.23 |
| 400 | 439.61 |
| 500 | 625.92 |
| 1000 | 1864.45 |

However, as displayed in Fig. 3, while performance improved dramatically when the sizes of smaller dictionaries were increased, the effect eventually reached a plateau. The optimum dictionary size seems to be around 200 words.

We also tested the effect of the number of random samples, again using mean curvature and a dictionary size of 500. Sample points were uniformly re-sampled when the number of desired samples changed. Like with dictionary size, again, we saw performance improvement reach a plateau as the number of sample points increased; a choice of around 600 sample points seems to be the best.

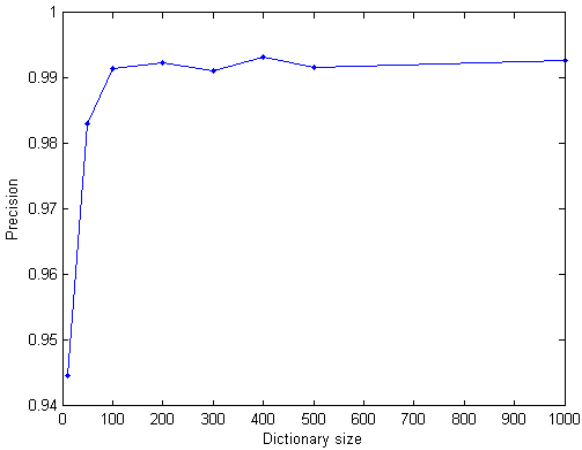

Figure 3. Effect of dictionary size on retrieval. Mean curvature was used as a local descriptor and 500 random points were found on each object.

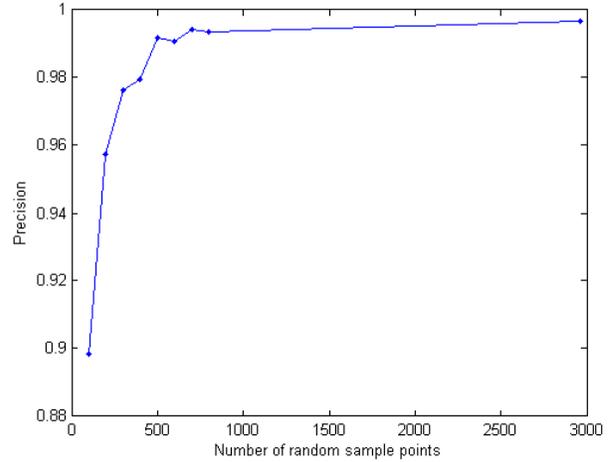

Figure 4. Effect of number of sample points on retrieval. Mean curvature was used as the local descriptor and the visual dictionary had 500 words.

### 6.3 Salient point detectors

**Determination of parameters for 3D-Harris rings**

Only two parameters improved the performance of points detected by 3D-Harris when returned – changing the neighborhood sampling to rings and increasing the fraction of points selected. We used mean curvature as the local descriptor, with a dictionary of 500 words. Figure 5 and Figure 6 display the effect of the relevant parameters.

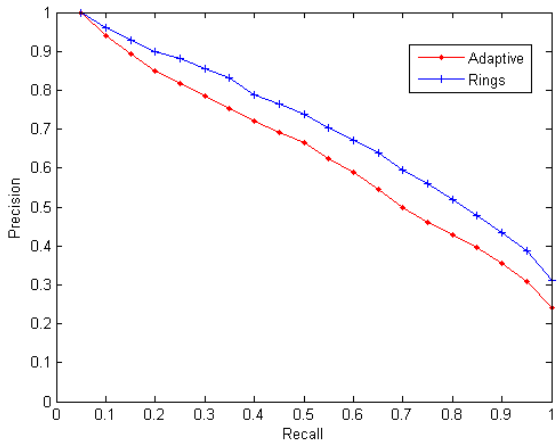

Figure 5. Precision vs. recall curves using sample points returned by the 3D-Harris algorithm using different methods to find vertex neighborhoods. Mean curvature was the local descriptor and the visual dictionary had 500 words.

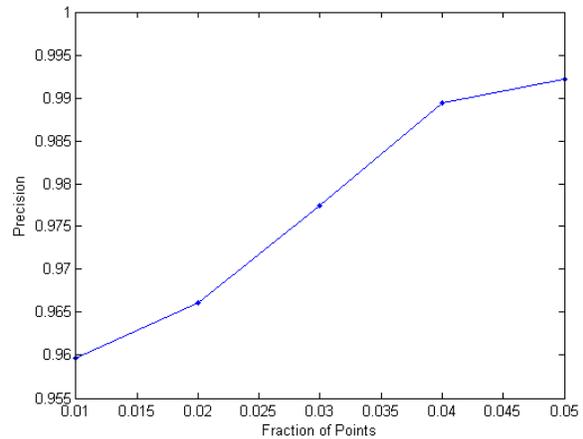

Figure 6. Effect of selecting different fractions of points to be returned by 3D-Harris as sample points on retrieval. Neighborhoods were selected using the rings method, mean curvature was the local descriptor, and the visual dictionary had 500 words.

It can be seen that using rings to find neighborhoods outperformed the adaptive scheme. In addition, selecting a larger fraction of points also improved performance. The final parameters used for 3D-Harris rings were ring neighborhoods with a size of one ring, $k = 0.01$, where $k$ is the constant in (6), $r = 1$, where $r$ is the number of rings considered a part of a vertex's neighborhood when determining local maxima, and 5 % of points selected as salient points.

**Comparing salient point detectors**

The different points detected by each salient point detection method on the armadillo model – file T261.off in the SHREC 2011 Non-Rigid 3D Watertight Meshes Dataset – are shown in Fig. 7.

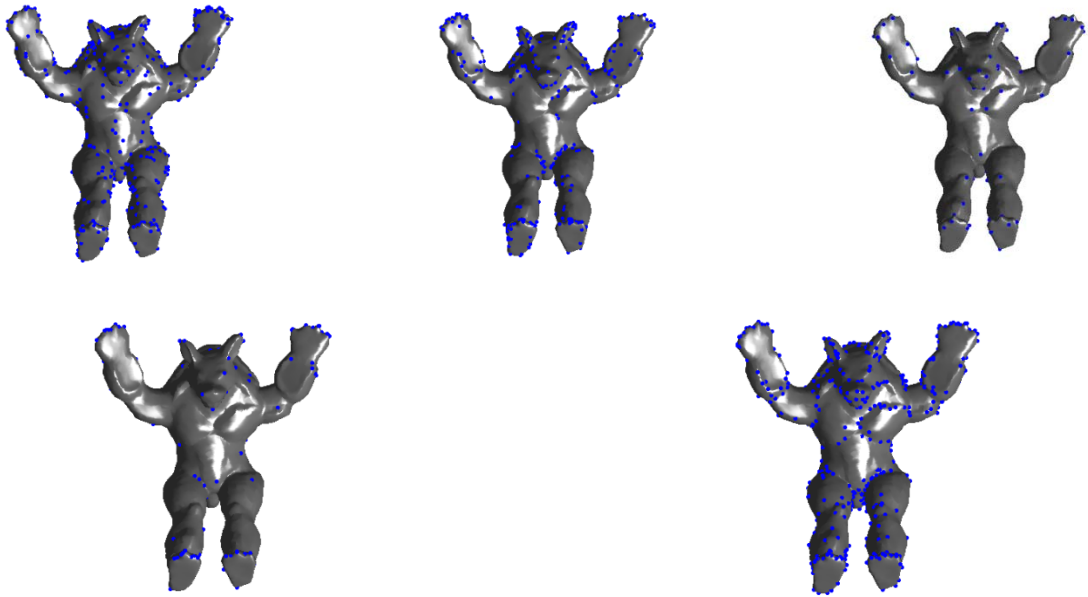

Figure 7. Salient point detected. Top row (left to right): 500 random points, mesh saliency, salient points. Bottom row (left to right): 3D-Harris adaptive, 3D-Harris rings

All methods detected salient points in similar areas of the model – toes, fingers, ears, and creases. Especially notable differences were seen in the number of points on smoother areas of the mesh; mesh saliency and 3D-Harris rings returned more points on the chest, arms, and legs than salient points and 3D-Harris adaptive. 3D-Harris rings returned a similar density of points to 500 random points; however, these points were clearly clustered more towards the creases of the mesh.

To test performance with the bag-of-words algorithm, we sampled the mean curvature at the vertices returned by each salient point detector. We used a dictionary of 500 words. Table 3 and Fig. 8 show our results.

For most recall values, using 500 random points still had the highest precision. 3D-Harris rings was a close second, followed by mesh saliency. However, at high recall values (greater than 0.8), mesh saliency performed the best. Salient points and 3D-Harris adaptive performed the worst.

From Table 3, we see that methods returning more points generally performed better, suggesting that the effectiveness of a salient point detector was determined at least in part by the number of points it returned.

When the number of sample points chosen was small, sampling salient points instead of random ones was able to improve performance. Both salient points and 3D-Harris adaptive clearly gave better results than sampling 100 random points, while both methods involved fewer sample points. It is also interesting to note that while 3D-Harris adaptive returned more points on average and in total, it still did not perform as well as salient points. Looking at Fig. 7, one explanation might be that 3D-Harris adaptive returned more points on the right of the armadillo than the left, while salient points returned a more symmetrical distribution of points. 3D-Harris adaptive also returned virtually no points on the armadillo's body, arms, or legs, while salient points has sample points representing features like the chest and the knees. In this case, calculating a local descriptor at salient points was able to provide more information and improve performance.

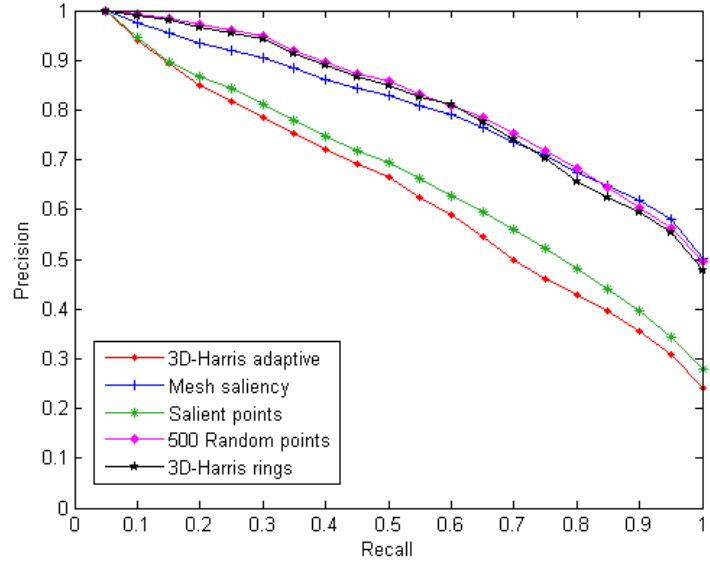

Figure 8. Precision vs. recall curves for sample point selection methods. Mean curvature was the local descriptor and the visual dictionary had 500 words.

As the number of sample points increased however, a continued performance improvement did not occur. For example, mesh saliency gave approximately the same statistics as using 300 or 400 random points. 3D-Harris rings and 500 random points, while they had different visual distributions of points, were also fairly equal in performance. This suggests that the benefit of detecting salient points also reaches a plateau, similar to what we saw with the dictionary's size and number of random sample points. At a larger number of sample points, sampling a local descriptor at salient points can slightly decrease the number of sample points needed to achieve a similar level of performance (as with 500 random points and 3D-Harris rings), but this advantage is not much.

Table 3. Retrieval statistics for sample point selection methods. Mean curvature was the local descriptor and the visual dictionary had 500 words.

|  | NN | 1-tier | 2-tier | E-measure | DCG | Avg. points | Total points |
|---|---|---|---|---|---|---|---|
| 500 random | 0.9850 | 0.7432 | 0.8480 | 0.6190 | 0.9281 | 500 | 300000 |
| 3D-Harris rings | 0.9817 | 0.7389 | 0.8355 | 0.6104 | 0.9235 | 461 | 276891 |
| 400 random | 0.9583 | 0.7272 | 0.8427 | 0.6127 | 0.9194 | 400 | 240000 |
| 300 random | 0.9533 | 0.6922 | 0.8205 | 0.5938 | 0.9038 | 300 | 180000 |
| Mesh saliency | 0.9500 | 0.7254 | 0.8444 | 0.6129 | 0.9128 | 354 | 212185 |
| 200 random | 0.9050 | 0.6404 | 0.7822 | 0.5601 | 0.8761 | 200 | 120000 |
| Salient points | 0.8900 | 0.5861 | 0.7445 | 0.5302 | 0.8452 | 75 | 44884 |
| 3D-Harris adaptive | 0.8767 | 0.5595 | 0.7073 | 0.5016 | 0.8287 | 93 | 55585 |
| 100 random | 0.7967 | 0.5131 | 0.6584 | 0.4659 | 0.7920 | 100 | 60000 |

### 6.4 Combining descriptors

First, 500 sample points were chosen for each object. Each descriptor was evaluated at these sample points and tested using a dictionary of 500 words as reference. We combined descriptors in four different ways. First, two descriptors were combined by concatenating the vectors returned when they were sampled at the same points on each object. Data from two descriptors combined this way are labeled with format *descriptor1+descriptor2* – VS. Alternatively, two descriptors' histograms were concatenated, again using the same sample points; this data is labeled with the format *descriptor1+descriptor2 – HistS*.

We also tried to not only sample two different descriptors, but also different points as well. Thus, we calculated the performance of two descriptors separately, choosing different sample points for each descriptor. The histograms were directly concatenated, and performance data is labeled with the format *descriptor1+descriptor2 – HistD*. Then, the vectors from each descriptor were concatenated and these combined vectors were tested as well; results are labeled with the format *descriptor1+descriptor2 – VD*. In addition, to see how well each descriptor corresponded with itself, we also sampled each descriptor at two sets of 500 sample points each and concatenated the resulting vectors (labeling results with format *descirptor*VD) and histograms (labeling results with format *descirptor*HistD).

As shown in Table 4, concatenating the vectors from calculating normal distribution and shape index at the same sample points performed the best. In general, some descriptors clearly corresponded better with each other, and only some of the combined descriptors performed better than mean curvature, the single local descriptor with the highest statistics. It is also interesting to note that neither normal distribution nor shape index performed the best as an individual descriptor, but together was the best combination. On the other hand, combining the mean curvature with some descriptors actually made performance worse than just mean curvature alone, including combining the descriptor with itself.

Concatenating vectors from different sample points clearly gave the lowest results. The top four results used two descriptors at the same sample points rather than different ones. In terms of vectors concatenation, using the same points consistently gave much better results. However, though the best combination came from concatenating vectors, concatenating histograms gave better performance overall. It is unclear whether using the same or different sample points is better when concatenating histograms – for example, using the same sample points performs better when combining mean curvature without another descriptor, yet using difference points is more preferable when combining normal distribution with others.

Table 4. Retrieval statistics for combinations of local descriptors.

|  | NN | 1-Tier | 2-Tier | e-Measure | DCG |
|---|---|---|---|---|---|
| ND+SI – VS | 0.9950 | 0.7978 | 0.8910 | 0.6513 | 0.9484 |
| Mean+SI – HistS | 0.9933 | 0.8106 | 0.9053 | 0.6603 | 0.9579 |
| Mean+CI – VS | 0.9917 | 0.7773 | 0.8724 | 0.6376 | 0.9419 |
| Mean+Gauss – HistS | 0.9917 | 0.7605 | 0.8487 | 0.6187 | 0.9344 |
| ND+SI – HistD | 0.9900 | 0.8240 | 0.9171 | 0.6701 | 0.9591 |
| Mean+SI – VS | 0.9900 | 0.7791 | 0.8793 | 0.6424 | 0.9447 |
| CI+SI – HistS | 0.9900 | 0.8253 | 0.9125 | 0.6680 | 0.9580 |
| ND+Gauss – HistD | 0.9900 | 0.7649 | 0.8500 | 0.6224 | 0.9312 |
| SI+Gauss – Hist D | 0.9900 | 0.7973 | 0.8922 | 0.6520 | 0.9512 |
| Mean+Gauss – HistD | 0.9900 | 0.7655 | 0.8515 | 0.6220 | 0.9354 |
| SIHistD | 0.9900 | 0.7884 | 0.8989 | 0.6550 | 0.9519 |
| CI+SI – VS | 0.9883 | 0.8065 | 0.8963 | 0.6558 | 0.9526 |
| SI+DTP – HistD | 0.9883 | 0.7861 | 0.8980 | 0.6537 | 0.9489 |
| Mean+CI – HistS | 0.9867 | 0.7746 | 0.8637 | 0.6323 | 0.9385 |
| CI+SI – HistD | 0.9867 | 0.8215 | 0.9120 | 0.6656 | 0.9574 |
| CI+Gauss – HistD | 0.9867 | 0.7639 | 0.8502 | 0.6208 | 0.9330 |
| ND+SI – HistS | 0.9850 | 0.8241 | 0.9146 | 0.6682 | 0.9581 |
| Mean+SI – HistD | 0.9850 | 0.8161 | 0.9093 | 0.6637 | 0.9581 |
| Mean+CI – HistD | 0.9850 | 0.7777 | 0.8629 | 0.6324 | 0.9386 |
| Mean+ND – HistS | 0.9850 | 0.7816 | 0.8647 | 0.6344 | 0.9379 |
| ND+CI – HistD | 0.9850 | 0.7748 | 0.8585 | 0.6292 | 0.9347 |
| ND+CI – HistS | 0.9850 | 0.7704 | 0.8590 | 0.6291 | 0.9327 |
| CIHistD | 0.9850 | 0.7587 | 0.8529 | 0.6240 | 0.9300 |
| ND+Gauss – VS | 0.9833 | 0.7552 | 0.8487 | 0.6186 | 0.9295 |
| CI+Gauss – VS | 0.9833 | 0.7544 | 0.8533 | 0.6208 | 0.9291 |
| Mean+ND – HistD | 0.9833 | 0.7839 | 0.8661 | 0.6345 | 0.9394 |
| CI+DTP – HistS | 0.9833 | 0.7500 | 0.8588 | 0.6254 | 0.9302 |
| **Mean** | **0.9833** | **0.7448** | **0.8450** | **0.6178** | **0.9288** |
| SI+Gauss – VS | 0.9817 | 0.7740 | 0.8685 | 0.6351 | 0.9377 |
| Mean+Gauss – VS | 0.9817 | 0.7833 | 0.8716 | 0.6367 | 0.9407 |
| ND+CI – VS | 0.9817 | 0.7582 | 0.8538 | 0.6252 | 0.9296 |
| CI+DTP – HistD | 0.9817 | 0.7476 | 0.8544 | 0.6224 | 0.9292 |
| Mean+DTP – HistD | 0.9817 | 0.7573 | 0.8644 | 0.6307 | 0.9356 |
| NDHistD | 0.9817 | 0.7632 | 0.8517 | 0.6247 | 0.9292 |
| SI+DTP – HistS | 0.9800 | 0.7858 | 0.9016 | 0.6566 | 0.9477 |
| Mean+DTP – HistS | 0.9800 | 0.7593 | 0.8632 | 0.6307 | 0.9346 |
| MeanHistD | 0.9800 | 0.7395 | 0.8477 | 0.6173 | 0.9269 |
| SI+DTP – VS | 0.9783 | 0.7468 | 0.8756 | 0.6342 | 0.9344 |
| CI+Gauss – HistS | 0.9783 | 0.7640 | 0.8508 | 0.6204 | 0.9318 |
| Mean+DTP – VS | 0.9783 | 0.7335 | 0.8531 | 0.6173 | 0.9269 |
| SI+Gauss – VD | 0.9767 | 0.7818 | 0.8767 | 0.6398 | 0.9421 |
| Mean+ND – VS | 0.9767 | 0.7734 | 0.8634 | 0.6321 | 0.9350 |
| ND+DTP – HistD | 0.9733 | 0.7452 | 0.8543 | 0.6226 | 0.9278 |
| **CI** | **0.9733** | **0.7405** | **0.8475** | **0.6173** | **0.9231** |
| Mean+ND – VD | 0.9717 | 0.7091 | 0.8333 | 0.6057 | 0.9125 |
| Gauss+DTP – HistD | 0.9717 | 0.7210 | 0.8320 | 0.6041 | 0.9192 |
| GaussHistD | 0.9717 | 0.6922 | 0.8016 | 0.5805 | 0.9069 |
| **SI** | **0.9700** | **0.7441** | **0.8733** | **0.6335** | **0.9336** |
| Mean+CI – VD | 0.9667 | 0.7180 | 0.8415 | 0.6104 | 0.9148 |
| CI+DTP – VS | 0.9667 | 0.7225 | 0.8382 | 0.6097 | 0.9184 |
| Gauss+DTP – HistS | 0.9650 | 0.7182 | 0.8323 | 0.6032 | 0.9177 |
| ND+DTP – HistS | 0.9633 | 0.7443 | 0.8541 | 0.6223 | 0.9255 |
| Gauss+DTP – VS | 0.9633 | 0.7019 | 0.8273 | 0.5975 | 0.9123 |
| CI+SI – VD | 0.9600 | 0.7349 | 0.8625 | 0.6237 | 0.9238 |
| **ND** | **0.9600** | **0.7351** | **0.8451** | **0.6154** | **0.9171** |
| ND+Gauss – VD | 0.9583 | 0.6958 | 0.8221 | 0.5956 | 0.9036 |
| ND+CI – VD | 0.9583 | 0.6958 | 0.8221 | 0.5956 | 0.9036 |
| CIVD | 0.9583 | 0.6910 | 0.8178 | 0.5942 | 0.9002 |
| NDVD | 0.9567 | 0.6668 | 0.8000 | 0.5773 | 0.8913 |
| ND+SI – VD | 0.9550 | 0.7182 | 0.8537 | 0.6186 | 0.9210 |
| Mean+SI – VD | 0.9533 | 0.7089 | 0.8556 | 0.6159 | 0.9143 |
| ND+DTP – VS | 0.9533 | 0.6740 | 0.8132 | 0.5873 | 0.8978 |
| CI+Gauss – VD | 0.9417 | 0.6645 | 0.7900 | 0.5714 | 0.8877 |
| MeanVD | 0.9400 | 0.6877 | 0.8238 | 0.5946 | 0.8984 |
| ND+Gauss – HistS | 0.9367 | 0.6474 | 0.7732 | 0.5580 | 0.8840 |
| SI+Gauss – HistS | 0.9367 | 0.6474 | 0.7732 | 0.5580 | 0.8840 |
| **Gauss** | **0.9367** | **0.6474** | **0.7732** | **0.5580** | **0.8840** |
| DTPHistD | 0.9333 | 0.6416 | 0.7897 | 0.5649 | 0.8846 |
| Mean+Gauss – VD | 0.9300 | 0.6728 | 0.8096 | 0.5812 | 0.8939 |
| SI+DTP – VD | 0.9283 | 0.6595 | 0.8282 | 0.5914 | 0.8869 |
| SIVD | 0.9150 | 0.6263 | 0.7905 | 0.5645 | 0.8704 |
| Mean+DTP – VD | 0.9133 | 0.6494 | 0.8068 | 0.5777 | 0.8815 |
| ND+DTP – VD | 0.9017 | 0.6121 | 0.7673 | 0.5478 | 0.8604 |
| **DTP** | **0.9000** | **0.6073** | **0.7616** | **0.5422** | **0.8629** |
| Gauss+DTP – VD | 0.8900 | 0.6024 | 0.7496 | 0.5348 | 0.8541 |
| GaussVD | 0.8183 | 0.5303 | 0.6810 | 0.4829 | 0.8066 |
| DTPVD | 0.7983 | 0.5319 | 0.6983 | 0.4910 | 0.8040 |

## 7. CONCLUSION

We explored the capabilities of the bag-of-words algorithm through testing proposed local descriptors, algorithm parameters, sampling methods, and combining descriptors. We have found that mean curvature, shape index, and curvature index are the best descriptors. In addition, we have determined that there is a significant performance gain when increasing dictionary size and the number of random sample points from small values. Salient point detection methods are still limited by the number of sample points they return, and seem to also only significantly improve performance when the number of sample points is small. Overall, 3D-Harris rings and mesh saliency are the best methods. Finally, normal distribution and shape index, through concatenating the vectors returned from calculating each at the same sample points, gives the best performance. We believe these observations we present can be useful to future experiments.